\documentclass[10pt,twocolumn,letterpaper]{article}

\usepackage[pagenumbers]{cvpr} 

\definecolor{cvprblue}{rgb}{0.21,0.49,0.74}
\usepackage[pagebackref,breaklinks,colorlinks,allcolors=cvprblue]
{hyperref}
\usepackage{algorithm}  
\usepackage{algpseudocode} 
\usepackage[table,xcdraw]{xcolor}
\usepackage{multirow}
\usepackage{placeins}
\usepackage{float}

\def\confName{CVPR}
\def\confYear{2026}

\title{Bridging the 2D-3D Gap: A Hierarchical Semantic-Geometric Map \\ for Vision Language Navigation}

\author{%
  \begin{tabular}{cccc}
    Kailing Li\textsuperscript{1} &
    Tianwen Qian\textsuperscript{1, \dag} &
    Lijin Yang\textsuperscript{2} &
    Yuqian Fu\textsuperscript{3}
  \end{tabular} \\
  \begin{tabular}{ccc}
    Jingyu Gong\textsuperscript{1} &
    Xiaoling Wang\textsuperscript{1, \dag} &
    Liang He\textsuperscript{1}
  \end{tabular}\\[0ex]
  \textsuperscript{1}School of Computer Science and Technology, East China Normal University \\
  \textsuperscript{2}Bosch Corporate Research, Shanghai, China  \\
  \textsuperscript{3}King Abdullah University of Science and Technology \\
  \small{\texttt{51275901046@stu.ecnu.edu.cn, twqian@cs.ecnu.edu.cn}}
}

\begin{document}
\maketitle

{
  \renewcommand{\thefootnote}{} 
  \footnotetext{\textsuperscript{\dag}Corresponding author.}
}

\begin{abstract}
Vision-Language Navigation (VLN) enables embodied agents to reach target locations in unseen environments by following language instructions. Despite recent progress with vision-language models (VLMs), a critical semantic–geometric gap remains: while VLMs excel at language and 2D visual understanding, they struggle with 3D spatial reasoning and fail to capture the causal dynamics between actions and spatial transitions, resulting in unreliable navigation, particularly in zero-shot settings. To bridge this gap, we propose a Hierarchical Semantic–Geometric Map (HSGM) that transforms 3D geometric information into a structured representation compatible with VLMs, effectively linking them to the physical world. Specifically, HSGM is represented as a multi-channel top-down map organized into three levels: (1) geometric level that records navigable regions and obstacles, (2) semantic level that represents objects and their relations, and (3) decision level that supports high-level task reasoning and goal selection. During navigation, the VLM acts as a high-level semantic planner, interpreting the spatial layout encoded in the HSGM to select geometrically valid waypoints, while low-level, collision-free movements between waypoints are executed by a classical path-planning algorithm, fully decoupling semantic reasoning from action execution. Additionally, complex instructions are decomposed into subtasks to alleviate the problem of progress forgetting or hallucinating in long-horizon navigation. Extensive experiments on R2R-CE and RxR-CE benchmarks demonstrate that our zero-shot framework achieves state-of-the-art performance and even outperforms several supervised methods. Code is available at \url{https://github.com/Teacher-Tom/HSGM_public}.
\end{abstract}    
\section{Introduction}
\label{sec:intro}

Vision-Language Navigation (VLN)~\cite{anderson2018vision} aims to enable an agent to follow natural language instructions and reach target locations in complex, unseen environments. This task serves as a cornerstone for embodied AI, integrating visual perception, language understanding, spatial reasoning, and sequential decision-making within a single framework. Recent advances in large vision-language models (VLMs)~\cite{li2023blip, zhang2024vision, zheng2025multimodal, fu2025objectrelator} have profoundly transformed the VLN landscape. Pretrained on massive image–text pairs, VLMs exhibit remarkable cross-modal alignment, extensive world knowledge, and robust commonsense reasoning, which substantially improve an agent’s semantic understanding of language instructions and visual inputs.

\begin{figure}
    \centering
    \includegraphics[width=0.95\linewidth]{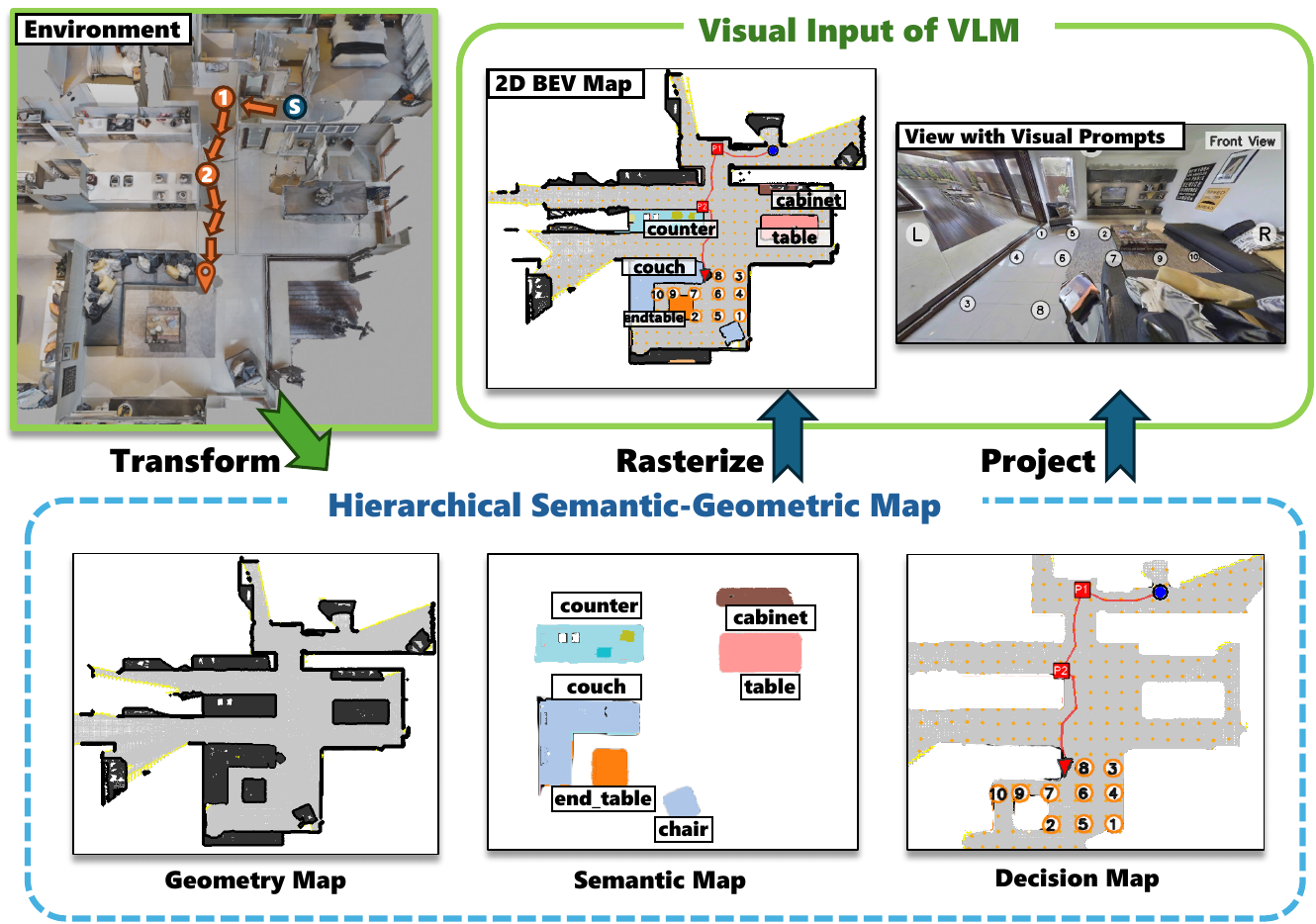}
    \caption{Our proposed \textbf{Hierarchical Semantic-Geometric Map (HSGM)}. 
The 3D environment is modeled via three maps: Geometry, Semantic, and Decision. 
It is then rasterized into a 2D BEV Map and projected as visual prompts onto the agent's view, serving as the structured visual input for the VLM.}
    \label{fig:concept}
\end{figure}
Despite these promising abilities, current VLMs remain unreliable navigators. The fundamental challenge lies in a \textbf{Semantic–Geometric Gap}: while VLMs excel at semantic understanding, they remain geometrically naive. Trained primarily on image–text pairs, their reasoning is confined to the plane of appearance, with little grasp of the underlying 3D geometry or how spatial relations evolve through physical interaction. This gap is mainly reflected in two interrelated weaknesses: 1) \textbf{Inadequate Spatial Understanding}. VLMs can recognize objects and their relations (e.g., \textit{the chair is next to the table}), but they struggle to infer global spatial layout across multiple viewpoints \cite{qi2025gpt4scene, yu2025far}. Consequently, their understanding of continuous space is fragmented and spatially ambiguous, preventing reliable alignment between instructions and 3D positions (e.g., \textit{pass between ...}); 2) \textbf{Ineffective Motion Planning}. VLMs are capable of high-level semantic planning (e.g., \textit{walk down the corridor, then turn toward the sofa}), but are not well-suited for translating such plans into low-level, physically executable action sequences (e.g., \textit{turn $15^\circ$ left, move forward $0.5$ m}). Prior methods, such as MapNav~\cite{zhang2025mapnav} and AO-Planner~\cite{chen2025affordances}, either force VLMs to predict raw actions directly or plan path from 2D images. Both strategies entangle semantic reasoning with geometric execution, pushing VLMs beyond their capability boundary.

To alleviate the aforementioned gap, an effective solution must explicitly bridge the semantic reasoning capability of VLMs with the geometric information of the physical world, while decoupling high-level planning from low-level control. To this end, we propose the \textbf{H}ierarchical \textbf{S}emantic–\textbf{G}eometric \textbf{M}ap (\textbf{HSGM}), a novel, training-free VLN framework. The central idea of HSGM is to represent the environment in a form that is both geometrically grounded and semantically interpretable to a VLM. As shown in Fig.~\ref{fig:concept}, HSGM decomposes the environment into three hierarchical levels: (1) \textbf{Geometric Map}, capturing navigable regions, obstacles; (2) \textbf{Semantic Map}, identifying and labeling object instances; (3) \textbf{Decision Map}, maintaining candidate waypoints, subtask nodes, and historical trajectories. HSGM serves as an information hub guiding the VLM throughout both spatial understanding and decision-making. During spatial understanding, we rasterize the 3D HSGM into a top-down 2D semantic map. This conversion allows the model to reason over spatial layouts through its native 2D vision pipeline while retaining essential geometric cues. For decision-making, the framework decouples semantic planning from low-level control. The VLM acts solely as a high-level planner, selecting the next geometrically valid waypoint on the decision map, while all movements between waypoints are executed by the $A^*$ algorithm. Specifically, candidate waypoints are uniformly sampled from the navigable region and projected onto both the egocentric image and top-down map. This converts a continuous 3D planning problem into a discrete selection task that aligns well with the VLM’s reasoning capability. To further improve stability of long-horizon navigation, we introduce a subtask management mechanism that decomposes complex instructions into ordered, executable subtasks, mitigating progress forgetting and hallucination. In addition, explicit modeling of stairs enables our framework to handle multi-floor navigation in complex environments.

Comprehensive experiments on R2R-CE and RxR-CE benchmarks demonstrate that our zero-shot framework achieves state-of-the-art performance, with success rates of 47.9\% and 41.8\%, respectively, surpassing all existing zero-shot methods and even outperforming several supervised ones. Our contributions are summarized as follows:

\begin{itemize}
    \item \textbf{Hierarchical Scene Representation.} We propose HSGM, a Hierarchical Semantic-Geometric Map that effectively bridges VLMs with the physical world.
    \item \textbf{Decoupled Planning \& Novel Waypoint Sampling.} Our framework decouples high-level semantic planning from low-level control. Furthermore, our waypoints are sampled from the 3D geometric map, enabling flexible and precise navigation without additional training.
    \item \textbf{State-of-the-Art Performance.} HSGM achieves state-of-the-art results on R2R-CE and RxR-CE, surpassing all existing zero-shot methods.
\end{itemize}

\section{Related Work}
\label{sec:related_work}

\begin{figure*}[!ht]
    \centering
    \includegraphics[width=0.95\linewidth]{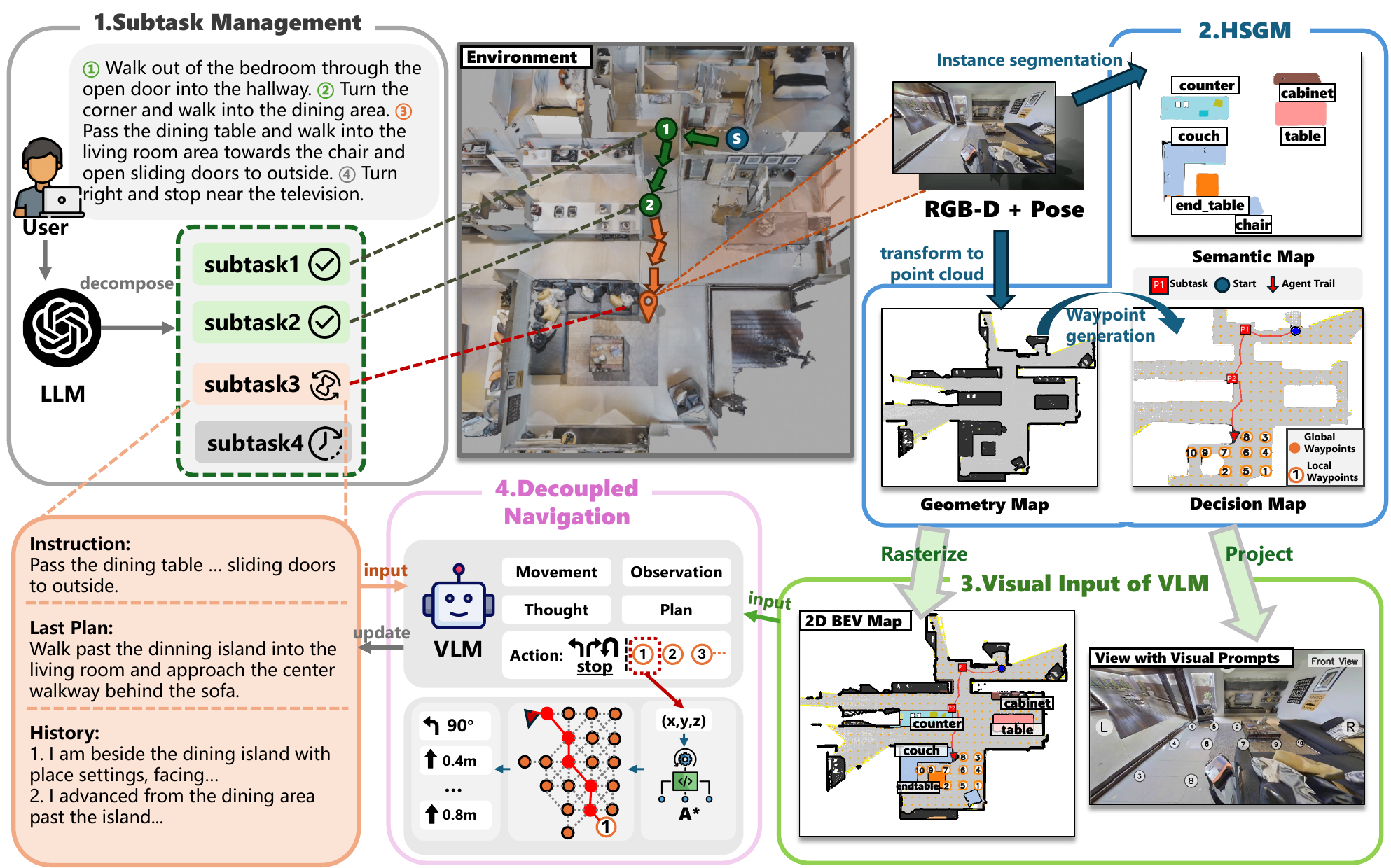}
    \caption{\textbf{Framework Overview.} (1) A LLM decomposes the user instruction into a sequence of subtasks. (2) The agent's sensor data (RGB-D, pose) is used to dynamically construct the 3D Hierarchical Semantic-Geometric Map. (3) The HSGM is rasterized into a 2D BEV map and projected onto the front view of the agent as visual input for the VLM. (4) The VLM performs CoT reasoning to select a waypoint, and the $A^*$ planner computes and executes a path to the selected waypoint according to the decision map.}
    \label{fig:framework}
\end{figure*}

\textbf{Vision-Language Navigation.}
VLN aims to enable an embodied agent to navigate in 3D environments via natural language instructions. Early VLN benchmarks, such as R2R~\cite{anderson2018vision}, modeled navigation as graph traversal in discrete environments, limiting real-world applicability~\cite{anderson2021sim}. The field has since shifted towards continuous environments (VLN-CE)~\cite{krantz2020beyond} with modern simulators like Habitat \cite{savva2019habitat}, where agents must execute low-level, physically feasible actions, necessitating geometric understanding and low-level control~\cite{hong2022bridging}.
As for the methods, traditional works rely on imitation or reinforcement learning~\cite{chen2022think, wang2023scaling, wang2024discovering}, demanding large-scale, domain-specific datasets (\emph{e.g.}, NaVid~\cite{zhang2024navid}). To overcome this, recent works explore zero-shot VLN using pre-trained VLMs (\emph{e.g.}, NavGPT~\cite{zhou2024navgpt}), though mostly in discrete settings. Our proposed HSGM tackles the more challenging and practical problem of zero-shot VLN-CE.

\noindent
\textbf{Scene Representation for VLN.}
Structured scene representations enhance the spatial understanding of VLN~\cite{wang2023gridmm, hong2023learning}. Some topological-based methods like MapGPT~\cite{chen2024mapgpt} and MC-GPT~\cite{zhan2024mc} convert graphs into text prompts for reasoning, but this abstraction sacrifices geometric fidelity. Other methods attempt to balance geometric fidelity with VLM accessibility. InstructNav~\cite{long2024instructnav} uses multi-sourced value maps for goal estimation while the map is invisible to the VLM. MapNav~\cite{zhang2025mapnav} introduces annotated 2D top-down maps as direct VLM inputs. Dynam3D~\cite{wang2025dynam3d} proposes a dynamic layered 3D representation to encode 3D spatial structures. However, these methods require in-domain training for the VLM to interpret such structured inputs. In contrast, our HSGM simultaneously leverages a 2D top-down semantic map for VLM-based spatial understanding and a 3D geometric map for precise, training-free route planning.

\noindent
\textbf{Action Policy for VLN.}
Action policies generally follow direct prediction or decoupled planning. Direct prediction methods~\cite{zhang2024navid, qi2025vlnr1} mainly train VLMs to output low-level actions, but suffer from data inefficiency. Zero-shot alternatives like AO-Planner~\cite{chen2025affordances} prompt VLMs for 2D affordance reasoning, which conflates visual and physical reachability, causing frequent collisions.
In contrast, decoupling methods are more robust. They can be divided into two categories: value-map-based methods~\cite{yokoyama2024vlfm, long2024instructnav, chen2025constraint} treat the VLM as a semantic scorer to guide a low-level planner, but rely heavily on perception quality~\cite{liu2024grounding}; waypoint-based methods~\cite{hong2022bridging, krantz2021waypoint, an2024etpnav} use trained predictors to generate sparse navigation waypoints (\emph{e.g.}, SmartWay~\cite{shi2025smartway}), but often yield inefficient movements between sparse points~\cite{wang2025dreamnav}. HSGM improves upon these designs by sampling dense, geometry-aware waypoints from a 3D map without training.


\section{Method}
\label{sec:method}

\subsection{Task Formulation}
\label{ssec:formula}

Vision-Language Navigation is formulated as a sequential decision-making problem. 
At each time step $t$, the agent receives a language instruction $I$, an egocentric observation $\overline{O_t}$, 
and predicts an action $a_t$ according to a navigation policy $\pi_\theta$. 
Formally, this process can be expressed as:
\begin{equation}
    a_t = \pi_\theta(I, \overline{O_t}, H_t),
\end{equation}
where $H_t = \{(\overline{O_k}, a_k)\}_{k < t}$ denotes the agent’s historical context, 
including all previous observations and actions. Observation $\overline{O_t}$ captures the agent’s egocentric perception of the environment at time $t$, typically consists of multi-view RGB-D images:
\begin{equation}
    \overline{O_t} = \{V_t^i\}_{i=1}^{3}, \quad
    V_t^i = \big(I_t^{i,\text{RGB}}, I_t^{i,\text{D}}\big),
\end{equation}
where $I_t^{i,\text{RGB}}$ and $I_t^{i,\text{D}}$ denote the RGB image and corresponding depth map from the $i$-th view (front, left, or right). The action $a_t \in \mathcal{A}$ is selected from a predefined action space that includes either low-level motion primitives
(e.g., \texttt{FORWARD}, \texttt{LEFT}, \texttt{RIGHT}, \texttt{STOP})
or high-level waypoints for planning. An episode terminates when the agent chooses the \texttt{STOP} action, and navigation is considered \emph{successful} if the final position $s_T$ lies within a threshold distance $\delta$ from the target location $s^*$.



\subsection{Framework Overview}
\label{ssec:framework}
As illustrated in Figure~\ref{fig:framework}, our framework's pipeline proceeds as follows. First, the natural language instruction $I$ input by the user is decomposed by an LLM (using the same VLM model as the navigation agent) into multiple subtasks $\mathcal{T} = \{T_1, \dots, T_k\}$, which include three states ($\{\texttt{done}, \texttt{pending}, \texttt{in\_progress}\}$). The navigation plan and historical trajectory descriptions generated during subtask execution are stored for use as the VLM's text input. Second, the agent executes each subtask sequentially, dynamically constructing the \textbf{Hierarchical Semantic-Geometric Map (HSGM)} in the process. This map consists of three layers: (1) a \textbf{Geometric Map ($M_{\text{geo}}$)}, constructed by converting sensor-derived RGBD images $\overline{O_t}$ and pose $\xi_t$ into a 3D geometric point cloud; (2) a \textbf{Semantic Map ($M_{\text{sem}}$)}, which combines YOLO-E~\cite{wang2025yoloe} for semantic instance segmentation on 2D RGB images and projects them into 3D space to obtain the location and class of instances in the scene; and (3) a \textbf{Decision Map ($\mathcal{M}_{\text{dec}}$)}. On the navigable area point cloud of the Geometric Map, we sample \emph{Global Waypoint Graph ($G$)} and a \emph{Local Waypoint Set ($A_{\text{curr}}$)}. Global waypoints provide a discrete topological map of the scene for low-level path planning, while local waypoints enable the VLM to make visualized, high-level semantic decisions. Furthermore, the Decision Map records the agent's historical trajectory $\tau_{\text{his}}$ and the completion locations of subtask nodes $\{\pi_{\text{done}, k}\}$. 

Then, the HSGM is rasterized into a 2D BEV map ($\mathcal{M}_{\text{bev}}$), and the local waypoints $A_{\text{curr}}$ are projected onto the first-person image ($V_t^{\text{front}}$) to form visual prompts; these two elements constitute the VLM's visual input.
Subsequently, based on the visual and text inputs, the VLM performs high-level semantic planning by outputting structured Chain-of-Thought (CoT) reasoning. This reasoning process includes \texttt{Movement}, \texttt{Observation}, \texttt{Thought}, \texttt{Plan}, and culminates in an \texttt{Action} selected from the high-level action space $A_t = A_{\text{turn}} \cup A_{\text{curr}}$, or \texttt{STOP}. Finally, after the VLM makes a decision, if a waypoint $g \in A_{\text{curr}}$ is selected, the A* algorithm is used to plan the optimal route to that location based on the waypoint's coordinates and the \emph{Global Waypoint Graph $G$}. This path is then converted into a low-level action sequence ($\pi_{\text{L}}$) for execution.

\subsection{Hierarchical Semantic-Geometric Map}
\label{ssec:hsgm}

HSGM serves as a crucial bridge between the VLM and the physical world. It maintains a dynamically updated 3D representation of the surrounding environment, which is organized into three complementary levels: a geometric map, a semantic map, and a decision map. In the following, we detail the definition and construction of each component.

\noindent
\textbf{Geometric Map} forms the spatial backbone of the HSGM, capturing the global topology and geometric layout of the environment.
Following InstructNav~\cite{long2024instructnav}, we back-project pixels of multi-view RGB images $I_{t}^{i,\text{RGB}}$ with depth maps $I_{t}^{i,\text{D}}$ and camera poses $\xi_t$ into 3D scene space to form a scene point cloud $P_{\text{scene}}$. Points located above the ground plane are defined as obstacles $P_{\text{obs}}$, and the remaining regions without obstacles are extracted as the initial traversable area, denoted as $P_{\text{nav}}^{\text{init}}$.
To support cross-floor navigation, we further detect stairs $P_{\text{stair}}$ via surface normal estimation and tilted-plane filtering, and incorporate them into the navigable set:
\begin{equation}
P_{\text{nav}} = P_{\text{nav}}^{\text{init}} \cup P_{\text{stair}}.
\end{equation}

The final geometric map aggregates both the navigable regions and obstacles:
\begin{equation}
M_{\text{geo}} = P_{\text{nav}} \cup P_{\text{obs}},
\end{equation}
which serves as the geometric foundation for subsequent motion planning and control.

\noindent
\textbf{Semantic Map} enables the VLM to perceive the scene content from a semantic perspective. 
To extract object-level semantics, we employ a YOLO-E~\cite{wang2025yoloe} instance segmentation model over the egocentric RGB images $I_{t}^{i,\text{RGB}}$ to obtain a set of 2D instance masks $\{M_j\}$ and their corresponding class labels $\{c_j\}$. Each mask $M_j$ is then back-projected into 3D space using the associated depth map $I_{t}^{i,\text{D}}$ and camera pose $\xi_t$, generating an instance-level point cloud $P_{\text{obj}, j}$. As the agent moves, $P_{\text{obj}, j}$ from multiple frames are merged when they exhibit high 3D IoU and consistent semantic labels, forming a temporally coherent instance-level semantic map. To suppress noise, instances with insufficient points are discarded.
The final semantic map $M_{\text{sem}}$ is defined as a collection of observed object instances:
\begin{equation}
    M_{\text{sem}} = \{ (P_{\text{obj}, j}, c_j) \}_{j=1}^{N_{\text{obj}}}.
\end{equation}

\begin{algorithm}[t]
\caption{Local Waypoint Generation ($A_{\text{curr}}$)}
\label{alg:waypoint_gen}
\begin{algorithmic}[1]
\State \textbf{Input:} $P_{\text{nav\_view}}$, $P_{\text{obs}}$, $p_{\text{agent}}$
\State \textbf{Output:} $A_{\text{curr}}$
\State $P_{\text{clean}} \gets \Call{Denoise}{P_{\text{nav\_view}}}$
\State $C_{\text{sparse}} \gets \Call{VoxelDownsample}{P_{\text{clean}}}$
\State $C_{\text{valid}} \gets \emptyset$
\For{each point $p_c \in C_{\text{sparse}}$}
    \If{$\Call{CylindricalCheck}{p_c, P_{\text{obs}}, r_{\text{agent}}, h_{\text{agent}}}$}
        \State $C_{\text{valid}} \gets C_{\text{valid}} \cup \{p_c\}$
    \EndIf
\EndFor
\State $A_{\text{curr}} \gets \emptyset$
\For{each point $p_v \in C_{\text{valid}}$}
    \State $d_{\text{agent}} \gets \Call{Distance}{p_v, p_{\text{agent}}}$
    \State $d_{\text{obs}} \gets \Call{NearestDistance}{p_v, P_{\text{obs}}}$
    \If{$d_{\text{agent}} < 0.3$ \textbf{or} $d_{\text{agent}} > 3.0$ \textbf{or} $d_{\text{obs}} < 0.3$ \textbf{or} $d_{\text{obs}} > 2.0$}
        \State \textbf{continue}
    \EndIf
    \If{$\Call{HasPath}{p_{\text{agent}}, p_v}$}
        \State $A_{\text{curr}} \gets A_{\text{curr}} \cup \{p_v\}$
    \EndIf
\EndFor
\State \Return $A_{\text{curr}}$
\end{algorithmic}
\end{algorithm}

\noindent
\textbf{Decision Map} discretizes the navigable space of $M_{\text{geo}}$ into a set of waypoints, providing the foundation for movement planning and control. It consists of two components: a \emph{Global Waypoint Graph} $G$ and a \emph{Local Waypoint Set} $A_{\text{curr}}$. Formally, the decision map is denoted as:
\begin{equation}
\mathcal{M}_{\text{dec}} = \{ G, A_{\text{curr}} \},
\end{equation}
providing both global connectivity of the environment and local candidates for the agent’s navigation policy.

The global graph $G = (V, E)$ represents the navigable space in the form of a structured graph, enabling precise and collision-free low-level movement via the A* algorithm (refer to Sec.~\ref{ssec:navigation}). Nodes $V$ are generated by first denoising $M_{\text{geo}}$ and then applying voxel downsampling to produce a set of candidate positions $A_{\text{glob}}$. Each candidate $p_c \in A_{\text{glob}}$ undergoes a cylindrical occupancy check against obstacles $P_{\text{obs}}$ to ensure that the agent can safely occupy the position. A point is valid if:
\begin{equation}
\label{eq:cyl_check}
P_{\text{obs}} \cap \text{Cyl}(p_c, r, h) = \emptyset,
\end{equation}
where $r$ and $h$ represent the radius and height of the agent, respectively. Edges $E$ are established between node pairs $(v_i, v_j)$ that satisfy both a distance constraint ($|v_i - v_j|_2 \le 1.0\text{ m}$) and a height-difference threshold ($|v_i^z - v_j^z| \le 0.3\text{ m}$) to allow traversal across uneven terrain, including stairs. For each potential edge, we interpolate points along the line segment and check that all points fall within $M_{\text{geo}}$ and do not intersect $P_{\text{obs}}$. The resulting graph forms a dense, collision-free environment topology.

The local waypoint set $A_{\text{curr}}$ defines the dynamic action candidates available to the VLM at each time step. It is generated using a similar process (denoising, voxel downsampling, cylindrical occupancy check) applied to the agent’s current view with a coarser resolution (e.g., 1.0 m). Candidate points are further filtered with heuristics: (1) a \emph{Distance Filter} to retain points within a practical range; (2) a \emph{Semantic Filter} to prioritize points near objects; and (3) a \emph{Reachability Filter} to discard points unreachable from the global graph $G$.

\noindent
\textbf{2D Map Rasterization and Waypoint Visualization.}
Since VLMs are primarily trained on 2D image-text pairs and struggle to directly process 3D point clouds, we rasterize the 3D HSGM into an agent-centric multi-channel 2D bird's-eye-view (BEV) map $\mathcal{M}_{\text{bev}}$. $\mathcal{M}_{\text{bev}}$ serves as the core visual input to the VLM, containing: (1) a geometric channel with obstacles from $P_{\text{obs}}$ (black) and navigable areas from $P_{\text{nav}}$ (grey); (2) a semantic channel drawing class-specific markers for object instances from $M_{\text{sem}}$ at their geometric centers; and (3) state annotations overlaying the agent's current position, historical trajectory $\tau_{\text{his}}$, and the endpoints of completed subtasks $\{\pi_{\text{done}, k}\}$. To facilitate spatially grounded reasoning and decision-making, we further mark the local waypoints in $A_{\text{curr}}$ with numerical indices and project them onto both the BEV map $\mathcal{M}_{\text{bev}}$ and the agent’s forward-facing view $V_{t}^{\text{front}}$. Visited and unvisited waypoints are marked with different colors (e.g., red and gray) to enhance spatial disambiguation for the VLM.

\subsection{Decoupled Navigation}
\label{ssec:navigation}

Our navigation strategy adopts a decoupled paradigm, where the VLM is responsible for high-level semantic reasoning, and a classical path-planning algorithm governs low-level movement control. In the following, we detail the workflow.

\noindent
\textbf{Subtask Management.}
VLMs are prone to omitting instructions, skipping key steps, or incorrectly assessing task completion during long-horizon navigation\cite{chen2025constraint, song2025towards}. To mitigate this, we introduce a subtask management mechanism.
Initially, the VLM decomposes the complex instruction $I$ into ordered, executable subtasks $\mathcal{T} = \{T_1, \dots, T_k\}$. 
Each subtask $T_i$ satisfies two constraints: (1) \emph{Clear Termination}: a verifiable end-state (e.g., ``leave the bedroom''); and (2) \emph{Bounded Complexity}: no more than three motion directives.
During execution, a finite-state progress manager enforces sequential completion by tracking each subtask's state $\mathcal{S} \in \{\texttt{pending}, \texttt{in\_progress}, \texttt{done}\}$. The VLM is fed the \texttt{in\_progress} subtask until it outputs a special \texttt{STOP} action, which updates $T_i$'s state to \texttt{done} and activates $T_{i+1}$. To prevent premature termination, a double-confirmation mechanism (requiring two consecutive \texttt{STOP} outputs) is applied to the final subtask.

Two auxiliary mechanisms enhance robustness: (1) \emph{History Recording}: Completed subtask positions $\pi_{\text{done}, i}$ are recorded on the decision map and projected onto the BEV map $\mathcal{M}_{\text{bev}}$. These spatial anchors serve as historical references, preventing redundant exploration. (2) \emph{Automatic Backtracking}: If a subtask's step limit is exceeded, the agent automatically backtracks to its initial location to reattempt execution from a known state.

\begin{table*}[ht]
\centering
\caption{\textbf{Navigation performance on R2R-CE and RxR-CE benchmarks.} 
``*'' indicates that this method requires partial reliance on the simulator's labeled data for training.
Best zero-shot results are marked in \textbf{bold}, and the second-best is \underline{underlined}.}
\vspace{-6pt}
\scalebox{0.85}{
\begin{tabular}{c|l|cccc|cccc}
\hline
\multirow{2}{*}{Settings} 
& \multirow{2}{*}{Method} 
& \multicolumn{4}{c|}{R2R-CE (Val-Unseen)} 
& \multicolumn{4}{c}{RxR-CE (Val-Unseen)} \\ 
\cmidrule(lr){3-6} \cmidrule(lr){7-10}
& 
& SR $\uparrow$ & SPL $\uparrow$ & NE $\downarrow$ & OSR $\uparrow$
& SR $\uparrow$ & SPL $\uparrow$ & NE $\downarrow$ & nDTW $\uparrow$ \\ 
\hline
\multirow{8}{*}{Supervised} 
& Sasra~\cite{irshad2022sasra} \textcolor{gray}{ICPR 2022}                  & 24.0  & 22.0  & 8.32 & --     & --    & --    & --    & --    \\
& Seq2Seq~\cite{krantz2020beyond}  \textcolor{gray}{ECCV 2020}              & 22.0  & 25.0  & 7.77 & 37.0   & 13.9  & 11.9  & 12.10 & 30.8  \\
& CMA~\cite{hong2022bridging} \textcolor{gray}{CVPR 2022}                    & 32.0  & 30.0  & 7.37 & 40.0   & --    & --    & --    & --    \\
& Navid~\cite{zhang2024navid}  \textcolor{gray}{RSS 2024}                & 37.4  & 35.9  & 5.47 & 49.1   & 23.8  & 21.2  & 8.41  & --    \\
& ETPNav~\cite{an2024etpnav} \textcolor{gray}{TPAMI 2024}                & 57.0  & 49.0  & 4.71 & 65.0   & 54.8  & 44.9  & 5.64  & 61.9  \\
& MapNav~\cite{zhang2025mapnav}  \textcolor{gray}{ACL 2025}               & 39.7  & 37.2  & 4.93 & 53.0   & 32.6  & 27.7  & 7.62  & 43.5  \\
& Dynam3D ~\cite{wang2025dynam3d}  \textcolor{gray}{NeurIPS 2025}             & 52.9  & 45.7  & 5.34 & 62.1   & --    & --    & --    & --    \\
\hline
\multirow{8}{*}{Zero-shot} 
& SmartWay$^*$~\cite{shi2025smartway} \textcolor{gray}{IROS 2025}           & 29.0  & 22.5  & 7.01 & \underline{51.0}   & --    & --    & --    & --    \\
& OpenNav$^*$~\cite{qiao2025open} \textcolor{gray}{ICRA 2025}           & 19.0  & 16.1  & \underline{6.70} & 23.0   & --    & --    & --    & --    \\
& A2Nav$^*$~\cite{chen20232} \textcolor{gray}{NeurIPS 2023}             & 23.0  & 11.1  & --   & --     & 16.8  & 6.3   & --    & --    \\
& InstructNav~\cite{long2024instructnav}  \textcolor{gray}{CoRL 2024}          & 31.0  & 24.0  & 6.89 & --     & --    & --    & --    & --    \\
& AO-Planner~\cite{chen2025affordances} \textcolor{gray}{AAAI 2025}            & 25.5  & 16.6  & 6.95 & 38.3   & \underline{22.4}  & \underline{15.1}  & 10.75 & \underline{33.1}  \\
& CA-Nav~\cite{chen2025constraint}  \textcolor{gray}{TPAMI 2025} & 25.3  & 10.8  & 7.58 & 48.0   & 19.0  & 6.0   & \underline{10.37} & 13.5  \\
& DreamNav~\cite{wang2025dreamnav}  \textcolor{gray}{arXiv 2025}             & \underline{32.8}  & \underline{28.9}  & 7.06 & 41.0   & --    & --    & --    & --    \\
& \cellcolor[HTML]{EFF6FF}HSGM (Ours) 
& \cellcolor[HTML]{EFF6FF}\textbf{47.9} & \cellcolor[HTML]{EFF6FF}\textbf{32.8} & \cellcolor[HTML]{EFF6FF}\textbf{5.42} & \cellcolor[HTML]{EFF6FF}\textbf{58.7} 
& \cellcolor[HTML]{EFF6FF}\textbf{41.8} & \cellcolor[HTML]{EFF6FF}\textbf{25.1} & \cellcolor[HTML]{EFF6FF}\textbf{7.43} & \cellcolor[HTML]{EFF6FF}\textbf{54.9} \\
\hline
\end{tabular}}
\label{tab:r2r_rxr}
\end{table*}

\noindent
\textbf{Action Space of VLM.}
To decouple the high-level reasoning and low-level control, we define a discrete and dynamically adaptive action space for the VLM at each time step. Specifically, the candidate action space $A_t$ comprises two components: (1) a set of fixed \emph{Turn Actions}, $A_{\text{turn}} = \{\texttt{LEFT}(90^\circ), \texttt{RIGHT}(90^\circ), \texttt{TURN AROUND}(180^\circ)\}$, enabling orientation adjustment; and (2) a set of dynamic \emph{Waypoint Actions}, $A_{\text{curr}} = \{a_1, a_2, \dots, a_K\}$, representing geometrically reachable positions within the agent’s current field of view. Each waypoint in $A_{\text{curr}}$ corresponds to a short-term navigation goal derived from the decision map, offering the VLM a discrete spatially grounded option. The generation of these waypoints is fully training-free and described in detail in Alg.~\ref{alg:waypoint_gen} and Sec.~\ref{ssec:hsgm}.

\noindent
\textbf{High-Level Semantic Planning.}
The high-level planning policy $\pi_{\text{H}}$ is undertaken by the VLM. At each timestep $t$, the VLM receives: (1) \emph{Visual Inputs}, including ${V_{t}^{i}}$ and the BEV map $\mathcal{M}_{\text{bev}}$, both marked with the candidate waypoints $A_{\text{curr}}$; and (2) \emph{Task Inputs}, which provide the current instruction along with the accumulated reasoning history.
Based on these inputs, the VLM performs a structured Chain-of-Thought (CoT) reasoning to select the optimal action $a_t$ from the action space $A_t = A_{\text{turn}} \cup A_{\text{curr}}$, or output \texttt{STOP} to mark subtask completion. The CoT reasoning process follows a deliberate cognitive sequence: analyzing past progress, perceiving the current environmental state, inferring the immediate sub-goal, planning a feasible navigation step, and finally selecting an appropriate action. The full CoT prompts and reasoning examples are provided in the supplementary material.

\noindent
\textbf{Low-Level Motion Control.}
Once the high-level VLM policy $\pi_{\text{H}}$ selects a waypoint $a_t \in A_{\text{curr}}$ as the target $g$, the low-level motion controller $\pi_{\text{L}}$ is invoked to compute a geometrically precise and collision-free trajectory.
Specifically, given the target $g$, we use the A* algorithm as our $\pi_{\text{L}}$ over the pre-constructed global graph $G = (V, E)$ (refer to Sec.~\ref{ssec:hsgm}) to determine the shortest path $\tau_{\text{path}}$ from the agent’s current position. The cost function of A* is defined as the Euclidean distance between connected nodes.
The resulting path $\tau_{\text{path}}$ is then decomposed into a sequence of primitive control actions. For each segment, the agent first performs a \texttt{ROTATE} operation to align its orientation with the next waypoint, followed by a \texttt{FORWARD} action to advance. This low-level motion control pipeline enables the agent to execute the high-level semantic intentions of the VLM with geometric precision.

\section{Experiments}
\label{sec:experiments}

\subsection{Experimental Setting}
\label{ssec:exp_setting}

\textbf{Dataset.}
We evaluate on two standard VLN-CE benchmarks: R2R-CE~\cite{krantz2020beyond} and RxR-CE~\cite{ku2020room}. \textbf{R2R-CE} adapts R2R~\cite{anderson2018vision} from discrete graphs to continuous navigation in Habitat~\cite{savva2019habitat}, using photorealistic Matterport3D~\cite{chang2017mp3d} scenes and requiring low-level control. \textbf{RxR-CE} ports the large-scale, multilingual RxR dataset to Habitat, presenting a more challenging benchmark with long-horizon, pose-aligned instructions. Following prior zero-shot work, we test on the full R2R-CE validation unseen split and 500 randomly sampled English episodes from the RxR-CE validation unseen split. Ablation studies use a 300-episode subset of R2R-CE val unseen to reduce API costs. We utilize the GPT-5~\cite{openai2025gpt5} API as the core VLM in all experiments.

\noindent
\textbf{Evaluation Metrics.}
We follow prior work~\cite{hong2022bridging, shi2025smartway} and adopt the standard metrics for VLN-CE: Navigation Error (NE), Oracle Success Rate (OSR), Success Rate (SR), Success weighted by Path Length (SPL), and normalized Dynamic Time Warping (nDTW).

\subsection{Main Results}
\label{ssec:exp_results}

Tab.~\ref{tab:r2r_rxr} reports a comprehensive comparison between HSGM and prior state-of-the-art zero-shot and supervised VLN methods on R2R-CE and RxR-CE benchmarks. Our zero-shot HSGM consistently and substantially outperforms all existing zero-shot competitors across all primary navigation metrics, and even surpasses several fully supervised approaches.

\begin{table}[htbp]
\centering
\caption{\textbf{Ablation study of HSGM map in the BEV map.} 
``\checkmark'' indicates the map is used; ``$\times$'' means it is disabled. All results are reported on R2R Val-Unseen.}
\vspace{-6pt}
\scalebox{0.9}{
\begin{tabular}{ccc|cc}
\hline
Geo. & Sem. & Dec. & SR $\uparrow$ & SPL $\uparrow$ \\
\hline
$\times$ & $\times$ & $\times$ & 46.0 & 30.1 \\
\checkmark & $\times$ & $\times$ & 47.3 & 31.8\\
\checkmark & \checkmark & $\times$ & 49.2 & 32.8\\
\checkmark & \checkmark & \checkmark & \textbf{51.0} & \textbf{33.7}\\
\hline
\end{tabular}
}

\label{tab:map_ablation}
\end{table}

\noindent
\textbf{R2R-CE.} 
HSGM achieves a SR of 47.9\% and an SPL of 32.8\%. Compared to the strongest zero-shot competitor DreamNav~\cite{wang2025dreamnav}, we achieve improvements of 15.1\% and 3.9\% in SR and SPL, respectively. Furthermore, HSGM achieves the lowest NE at 5.42m and the highest OSR of 58.7\%. Together, these numbers indicate two complementary strengths of our approach: (1) improved \emph{semantic alignment} (higher SR/OSR), and (2) more \emph{geometrically precise execution} (lower NE, higher SPL). We attribute these gains to the HSGM’s ability to (i) present the VLM with a globally consistent BEV representation for better spatial grounding, and (ii) constrain semantic choices to geometrically validated waypoints, which reduces erroneous, infeasible action selection.

\noindent
\textbf{RxR-CE.}
The performance gap is more pronounced on this challenging long-horizon benchmark. Our \textbf{41.8\%} SR nearly doubles the strongest zero-shot baseline, AO-Planner~\cite{chen2025affordances} (22.4\% SR). Critically, our nDTW of \textbf{54.9\%} dramatically outperforms its 33.1\%, signifying superior spatiotemporal alignment. This validates the efficacy of our subtask management mechanism.

\noindent
\textbf{Comparison to Supervised Methods.}
Most notably, our zero-shot HSGM framework outperforms several fully supervised methods. For instance, on R2R-CE, our 47.9\% SR surpasses supervised methods like MapNav~\cite{zhang2025mapnav} (39.7\%) and NaVid~\cite{zhang2024navid} (37.4\%). This trend holds on RxR-CE, where our 41.8\% SR exceeds MapNav's 32.6\% and NaVid's 23.8\%. These results confirm that our approach is a highly effective and generalizable paradigm, capable of rivaling methods that rely on extensive in-domain training data.

\begin{table}[t]
\centering
\caption{\textbf{Ablation study on key components of decoupled navigation strategy.} 
All variants are evaluated on R2R Val-Unseen.}
\vspace{-6pt}
\scalebox{0.9}{
\begin{tabular}{l|cccc}
\hline
Method & SR $\uparrow$ & SPL $\uparrow$ & NE $\downarrow$ & OSR $\uparrow$ \\
\hline
Full Model & \textbf{51.0} & \textbf{33.7} & \textbf{5.24} & \textbf{61.7} \\
w/o subtasks dec.        & 42.1 & 28.9 & 5.59  & 57.9 \\
w/o plan-control sep.               & 44.3 & 31.9 & 5.47  & 57.0 \\
w/o structured CoT  & 34.0 & 18.0 & 6.48  & 55.3 \\

\hline
\end{tabular}
}

\label{tab:component_ablation}
\end{table}

\subsection{Ablation Study}
\label{ssec:abl_study}
We conduct ablation studies on the val-unseen subset of R2R-CE (300 episodes) to validate the contributions of the core components of our proposed method.

\noindent
\textbf{Impact of HSGM Projection in the BEV Map.}
Tab.~\ref{tab:map_ablation} quantifies the incremental benefit of rasterizing each HSGM map into the 2D BEV input. 
The baseline (no BEV map) achieves 46.0\% SR. 
Adding the \textbf{Geometric Map} (spatial layout, obstacles) boosts SR to 47.3\% (+1.3\%), as the top-down view resolves spatial ambiguities. 
Building on this, the \textbf{Semantic Map} (object categories and locations) further increases SR to 49.2\% (+1.9\%), enabling direct grounding of instructions to map objects. 
Finally, adding the \textbf{Decision Map} (waypoints, history, subtask nodes) reaches 51.0\% SR (+1.8\%), providing critical context for long-term reasoning. 
These results confirm that each layer of the rasterized HSGM provides a cumulative benefit, validating our design.

\begin{figure}[t]
    \centering
    \includegraphics[width=0.85\linewidth]{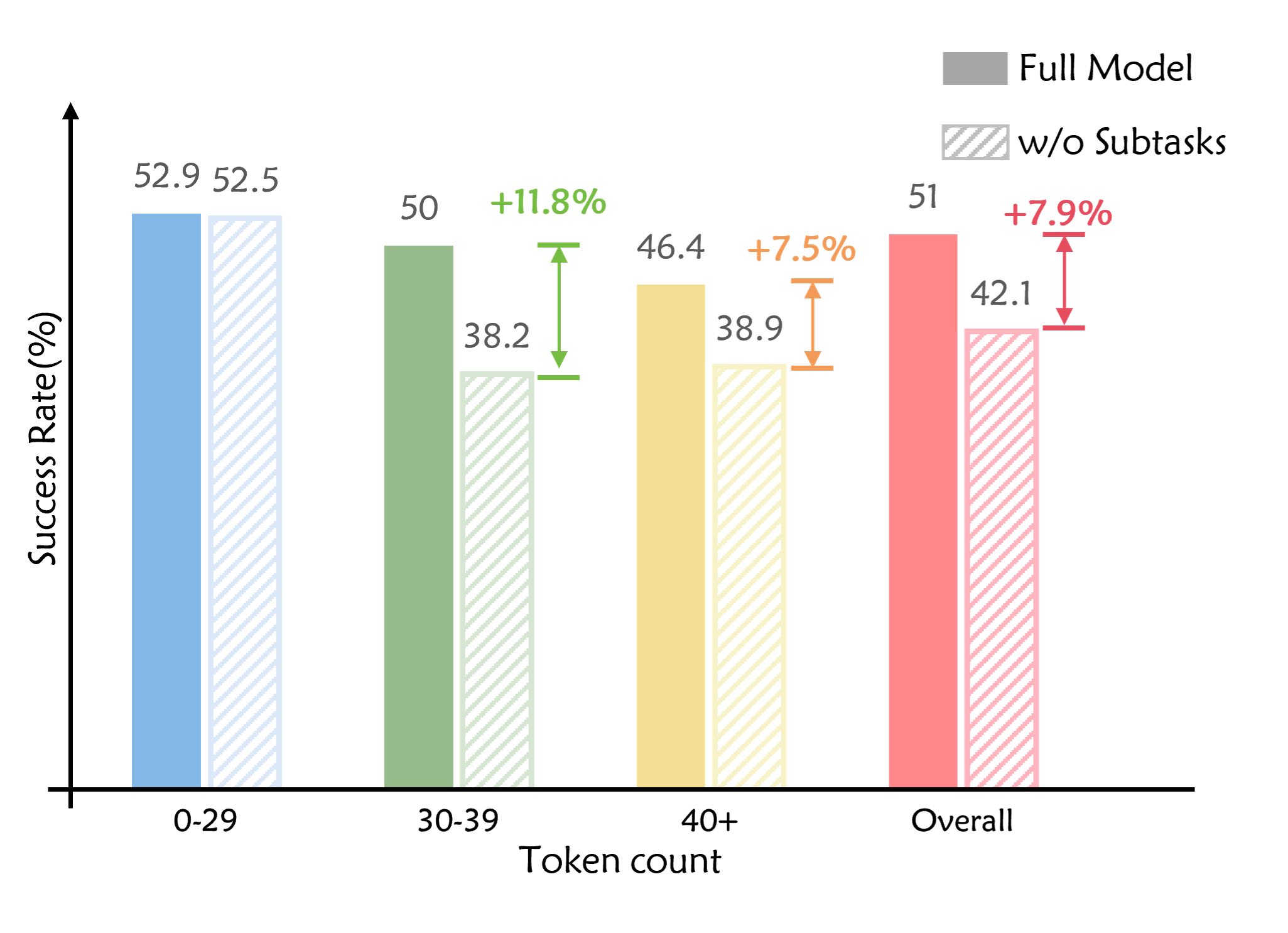}
    \vspace{-6pt}
    \caption{\textbf{Impact of Subtask Decomposition on Success Rate by Instruction Token Count.}}

    \label{fig:subtask_length}
\end{figure}

\begin{table}[t]
\centering
\caption{\textbf{Effectiveness of the Automatic Backtracking Mechanism.} The table shows the percentage of episodes where backtracking was triggered (Trigger Rate) and the success rate of those recovered episodes (Recovery SR).}
\vspace{-6pt}
\scalebox{0.85}{
\begin{tabular}{l|cc}
\hline
Benchmark & Trigger Rate & Recovery SR \\
\hline
R2R-CE(Val-unseen)    & 18.3\% & 30.8\% \\
RxR-CE(Val-unseen)    & 19.0\% & 26.8\% \\
\hline
\end{tabular}
}
\label{tab:backtracking}
\end{table}

\begin{figure*}[ht]
    \centering
    \includegraphics[width=0.9\linewidth]{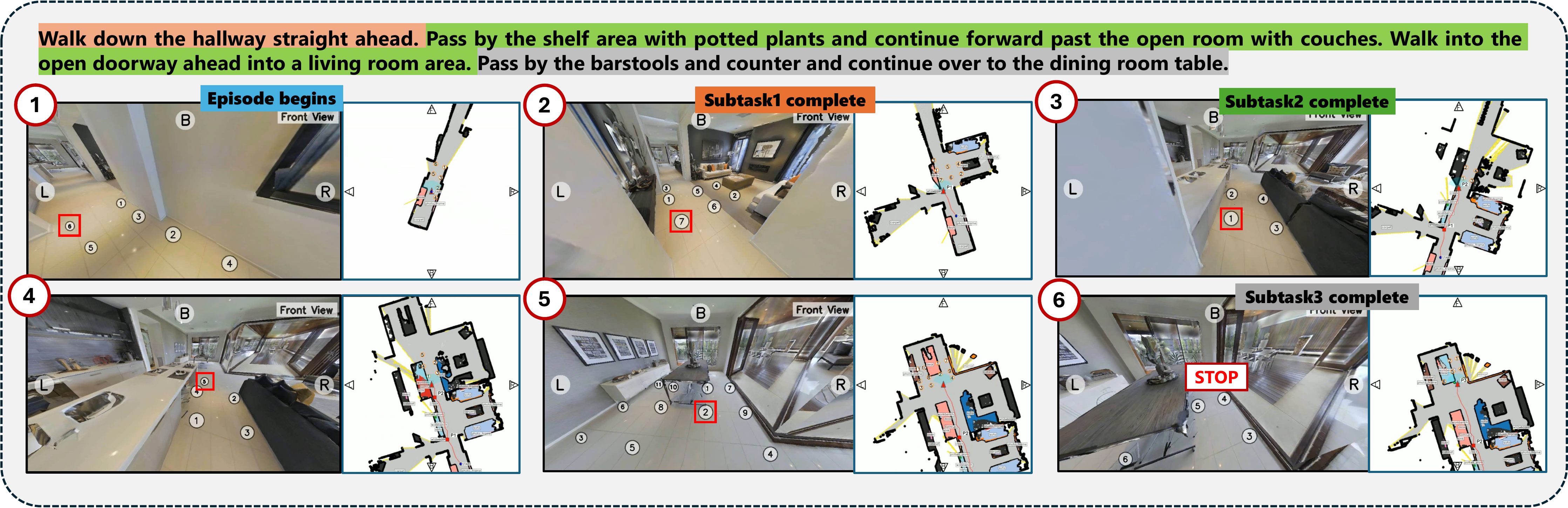}
    \caption{\textbf{Navigation visualization of HSGM}}

    \label{fig:case}
\end{figure*}

\noindent
\textbf{Impact of Decoupled Navigation Strategy.}
Tab.~\ref{tab:component_ablation} reports the ablation results of our decoupled navigation strategy, where the core mechanisms are removed from the full model: (1) Removing the \emph{subtasks decomposition mechanism} leads to a significant 8.9\% drop in SR. This degradation demonstrates that breaking long instructions into short, verifiable subtasks substantially reduces the VLM’s planning burden, thereby mitigating progress-forgetting issues. (2) Second, the ablation of \emph{planning-control separation} is implemented by replacing the $A^*$ planner with a simplified movement strategy of straight-line travel between waypoints, a strategy common to other waypoint-based methods like SmartWay~\cite{shi2025smartway}. As observed, this results in a 6.7\% decrease in SR. This indicates that high-level semantic planning must be predicated on reliable low-level motion control. (3) Third, the most deleterious ablation is the removal of the \emph{structured CoT} prompts. Without CoT prompting, SR plummets by 17.0\% to 34.0\%, and SPL also drops sharply by 15.7\%. This result confirms that VLN is indeed a complex reasoning task, it requires the agent not only to understand the 3D world but also to possess clear and logical thinking to effectively comprehend and execute tasks in physical environments.

\noindent
\textbf{Effect of Instruction Length on Subtask Decomposition.}
To further investigate how subtask decomposition mitigates the difficulty of long-horizon instructions, we partition episodes by instruction length and compare the full HSGM model against the \texttt{w/o subtasks dec.} variant, as shown in Fig.~\ref{fig:subtask_length}. For short instructions (with 0-29 tokens), full model and the variant perform comparably, indicating that succinct tasks can be handled without intermediate decomposition. As instruction length grows, however, a pronounced divergence appears.
In the medium-length (with 30–39 tokens), the variant suffers a steep SR drop to 38.2\%, while the full model retains 50.0\%, a gap of nearly 12\%. This deficit persists for the longest instructions ($\geq$40 tokens), where the non-decomposed policy continues to underperform significantly.
These results demonstrate that subtask decomposition substantially reduces the cognitive and memory burden of the VLM for long-horizon, multi-step episodes. By narrowing the immediate decision scope and providing verifiable termination signals, it prevents progress omission and hallucination.

\noindent
\textbf{Effectiveness of Automatic Backtracking.}
To further evaluate the robustness of our agent, we examine the contribution of the automatic backtracking mechanism.
As shown in Table~\ref{tab:backtracking}, backtracking was triggered in 18.3\% of R2R-CE episodes, successfully recovering 30.8\% of those failed trajectories. On the more challenging RxR-CE, it was activated in 19.0\% of episodes with a 26.8\% recovery rate. These results demonstrate that automatic backtracking plays a crucial role in enhancing task resilience, enabling the agent to recover from suboptimal navigation paths and avoid premature failure.

\subsection{Qualitative Results}
\label{ssec:qual_results}
Fig.~\ref{fig:case} illustrates a typical case study of our framework. First, the LLM decomposes the complex instruction into three sequential, verifiable subtasks. The agent then executes them in order. At each decision point, the VLM acts as a high-level semantic planner. It observes the HSGM-rasterized BEV map and the first-person view (which includes waypoint visual prompts) to select a high-level action. For instance, in step-1, it selects waypoint \texttt{6}. Subsequently, the $A^*$ algorithm computes and executes a precise, low-level action sequence to that waypoint.
When the VLM determines a subtask is complete, it outputs a \texttt{STOP} action. This completion is recorded on the BEV map (\emph{e.g.}, label \texttt{P1} for the first subtask), and the system proceeds to the next subtask. In step-6, the agent reaches the final goal (``dining room table'') and outputs \texttt{STOP} for the last subtask, successfully terminating the episode.

\section{Conclusions}

In this paper, we introduced the Hierarchical Semantic-Geometric Map (HSGM), a training-free framework that bridges the critical semantic-geometric gap in Vision-Language Navigation. HSGM transforms complex 3D environmental data into a structured representation, enabling a decoupled navigation paradigm. In our method, the VLM acts as a high-level semantic planner interpreting a 2D BEV map to make semantic decisions, while a classical $A^*$ algorithm executes robust low-level motion control. Our approach achieves state-of-the-art zero-shot performance on R2R-CE and RxR-CE benchmarks, outperforming all existing zero-shot and even several supervised methods. These results validate HSGM as a generalizable solution for grounding VLM reasoning in 3D geometric environments.

\section*{Acknowledgements}
\label{sec:ack}
This work was supported by Shanghai Municipal Science and Technology Major Project (No. 2025SHZDZX025G16), National Key R\&D Program of China (No. 2025ZD1801501), NSFC grant (No. 62136002 and 62477014), Ministry of Education Research Joint Fund Project (No. 8091B042239), Shanghai Knowledge Service Platform Project (No. ZF1213), and Shanghai Trusted Industry Internet Software Collaborative Innovation Center.
{
    \small
    \bibliographystyle{ieeenat_fullname}
    \bibliography{main}
}
\clearpage
\setcounter{page}{1}
\maketitlesupplementary

In this supplementary material, we provide additional details and comprehensive analyses to complement the main paper. 
Sec.~\ref{sec:impl} elaborates on the specific implementation details, covering the mechanism for multi-floor navigation, simulator configurations, and algorithmic hyperparameters. 
Sec.~\ref{supp_res} presents extended quantitative results, including the performance evaluation on the Object Goal Navigation task and a controlled comparative analysis to disentangle the impact of the VLM backbone from our framework design. 
Finally, Sec.~\ref{sec:more_vis} showcases more qualitative results, offering detailed visualizations of successful navigation trajectories and an in-depth analysis of typical failure cases.

\section{More Implementation Details}
\label{sec:impl}

\subsection{Implementation of Multi-Floor Navigation}
\label{ssec:multifloor}

To enable robust navigation in complex, multi-stair environments, we extend the HSGM framework to manage independent map representations for each floor. Let $\mathcal{F} = \{f_0, \ldots, f_n\}$ denote the set of floors. For each floor $f_i$, the system maintains a dedicated map instance $\mathcal{M}^{(i)}$ containing the floor-specific scene point cloud $\mathcal{P}_{scene}^{(i)}$, navigable areas $\mathcal{P}_{nav}^{(i)}$, and obstacles $\mathcal{P}_{obs}^{(i)}$. The active floor map is dynamically updated according to the following mechanism.

\paragraph{Floor Transition Mechanism.}
Floor transitions are governed by a state machine that monitors the agent's height relative to the current floor. A floor switch is triggered only when two conditions are satisfied: (1) the agent is located on a valid \textit{platform} (e.g., a landing), and (2) a significant vertical displacement is detected.
Specifically, platforms are detected using RANSAC plane fitting on local points within a radius of $r=1.5$m. A candidate plane $\Pi$ is validated if it exhibits a near-vertical normal ($|n_z| \ge 0.95$) and sufficient support area.
Let $\Delta H = z_{curr} - h_{floor}^{last}$ denote the vertical displacement relative to the previous floor height. The floor switching condition is then formulated as:
\begin{equation}
    \text{Switch} = \mathbb{I}(\text{isPlatform}(\Pi_{curr})) \land \left(|\Delta H| \geq \alpha \cdot H_{floor}\right),
\end{equation}
where $H_{floor}$ is the estimated floor-to-ceiling height and $\alpha = 0.75$ is a threshold factor. 
Crucially, this transition triggers a synchronous update of the visual input: the 2D BEV map $\mathcal{M}_{\text{bev}}$ is immediately switched to the rasterized representation of the new floor $\mathcal{M}^{(f_{\text{new}})}$, ensuring the VLM perceives the correct spatial context for subsequent planning.

\paragraph{Staircase Modeling and Planning.}
Stairs are geometrically distinct from flat terrain, often resembling obstacles due to their slope. To facilitate traversal, we explicitly identify stair regions $\mathcal{P}_{stair}$ by filtering scene points based on surface normal inclination, retaining points where the vertical component $|n_z| \in [0.2, 0.7]$. These points are spatially clustered using DBSCAN to filter noise.
During low-level planning, a waypoint $\mathbf{w}$ with a cylindrical agent footprint $\mathcal{C}(\mathbf{w})$ is considered valid if it is either collision-free or located within a detected stair region:
\begin{equation}
    \text{isValid}(\mathbf{w}) = \left(\mathcal{C}(\mathbf{w}) \cap \mathcal{P}_{obs} = \emptyset\right) \lor \left(\mathcal{C}(\mathbf{w}) \cap \mathcal{P}_{stair} \neq \emptyset\right).
\end{equation}
This mechanism effectively exempts staircases from standard obstacle constraints, allowing the $A^*$ planner to generate continuous paths across different elevations while maintaining safety on flat ground.

\begin{table}[t]
    \centering
    \small
    \setlength{\tabcolsep}{6pt}
    \caption{\textbf{Performance comparison on Object Goal Navigation.} \textbf{Bold} denotes the best zero-shot result, and \underline{underline} denotes the second best.}
    \label{tab:object_nav}
    \begin{tabular}{l|c|cc}
    \toprule
    \textbf{Method} & \textbf{Zero-shot} & \textbf{SR} $\uparrow$ & \textbf{SPL} $\uparrow$ \\
    \midrule
    Navid~\cite{zhang2024navid} & $\times$ & 32.5 & 21.5 \\
    MapNav~\cite{zhang2025mapnav} & $\times$ & 34.6 & 25.6 \\
    Uni-Navid~\cite{zhang2024uni} & $\times$ & 73.7 & 37.1 \\
    \midrule
    vlfm~\cite{yokoyama2024vlfm} & $\checkmark$ & 63.6 & 32.5 \\
    PIVOT~\cite{nasiriany2024pivot} & $\checkmark$ & 24.6 & 10.6 \\
    InstructNav~\cite{long2024instructnav} & $\checkmark$ & 58.0 & 20.9 \\
    ApexNav~\cite{zhang2025apexnav} & $\checkmark$ & \textbf{76.2} & \textbf{38.0} \\
    \textbf{HSGM (Ours)} & $\checkmark$ & \underline{73.6} & \underline{36.3} \\
    \bottomrule
    \end{tabular}
\end{table}

\subsection{Additional Experimental Settings}
\label{ssec:addition}

\paragraph{Simulator Configuration.}
We conduct our experiments using the Habitat simulator. The agent is modeled with a physical height of 1.2 m. The onboard visual sensor is configured with a Horizontal Field of View (HFOV) of $135^{\circ}$ and a downward tilt angle (pitch) of $35^{\circ}$ to optimize ground visibility. All visual observations are rendered at a resolution of $480 \times 640$ pixels.

\begin{table*}[ht]
    \centering
    \small
    \setlength{\tabcolsep}{4pt}
    \caption{\textbf{Comprehensive Analysis of Backbone vs. Framework.} We compare AO-Planner and HSGM under different configurations. Even without the BEV map, our decoupled framework (using only egocentric views) drastically outperforms the GPT-5-powered AO-Planner, highlighting the superiority of our navigation paradigm.}
    \label{tab:comprehensive_comparison}
    \begin{tabular}{l|c|c|cccc}
    \toprule
    \textbf{Method} & \textbf{Input View} & \textbf{Backbone} & \textbf{SR} $\uparrow$ & \textbf{SPL} $\uparrow$ & \textbf{NE} $\downarrow$ & \textbf{OSR} $\uparrow$ \\
    \midrule
    AO-Planner~\cite{chen2025affordances} & Ego-only & GPT-4o + Gemini 1.5pro & 25.5 & 16.6 & 6.95 & 38.3 \\
    AO-Planner$^\dagger$ & Ego-only & \textbf{GPT-5} & 32.3 & 21.3 & 6.12 & 45.0 \\
    \midrule
    \textbf{HSGM (w/o BEV)} & Ego-only & \textbf{GPT-5} & 46.0 & 30.1 & 5.37 & 61.7 \\
    \textbf{HSGM (Full)} & \textbf{Ego + BEV} & \textbf{GPT-5} & \textbf{51.0} & \textbf{33.7} & \textbf{5.24} & \textbf{61.7} \\
    \textbf{HSGM (Full)} & \textbf{Ego + BEV} & \textbf{GPT-4o} & 41.3 & 30.1 & 6.76 & 51.0 \\
    \bottomrule
    \end{tabular}
    \vspace{0.5em}
    \\
    \footnotesize{$^\dagger$: Re-implemented using the GPT-5 API.}
\end{table*}

\paragraph{Algorithmic Hyperparameters.}
In the waypoint generation phase, we employ a cylindrical collision model for the agent with a height of 1.2 m and a radius of 0.2 m to ensure geometric feasibility. For the low-level motion controller, the $A^*$ path planning algorithm is constrained to a maximum of 100 iterations. The maximum trajectory length is capped at 200 steps per episode.

\paragraph{Hardware Specifications.}
All experiments were executed on a workstation equipped with 96 GB of RAM and a single NVIDIA GeForce RTX 4090 GPU.

\section{Supplementary Results}
\label{supp_res}

\subsection{Performance on Object Goal Navigation}
\label{subsec:object_nav}

To further assess the versatility and generalization capability of our framework beyond instruction-following, we evaluated HSGM on the Object Goal Navigation (ObjectNav) task using the challenging Habitat-Matterport 3D (HM3D)~\cite{ramakrishnan2habitat} dataset. In this setting, the complex narrative instructions are replaced with a standardized template: ``\textit{Navigate to the target object [object] and get as close to it as possible.}'' Following standard protocols, a navigation episode is considered successful if the agent stops within a Euclidean distance of \textbf{0.3m} from the target object.

We compare HSGM against several supervised and zero-shot navigation agents. The results are summarized in Table~\ref{tab:object_nav}.

\paragraph{Results and Analysis.} 
Our framework achieves a remarkable Success Rate (SR) of \textbf{73.6\%} and an SPL of \textbf{36.3\%}.
First, compared to other zero-shot baselines, HSGM significantly outperforms methods like PIVOT (24.6\% SR), InstructNav (58.0\% SR), and vlfm (63.6\% SR). This demonstrates that our hierarchical semantic-geometric map provides a more robust representation for object localization and path planning than pure frontier-based or heuristic approaches.
Second, while slightly outperformed by ApexNav (76.2\% SR), which is specifically optimized for this exploration task, HSGM remains highly competitive, securing the second-best performance among all zero-shot methods.
Most notably, our zero-shot approach performs on par with the best supervised method, Uni-Navid (73.7\% SR), and substantially surpasses earlier supervised methods like Navid (32.5\% SR) and MapNav (34.6\% SR). This result validates that our explicit 3D mapping and decoupled planning strategy can achieve human-level perception and planning capabilities on the HM3D scenes without requiring extensive domain-specific training data.

\subsection{Disentangling Model Capabilities from Methodological Contributions}
\label{sec:backbone_comparison}

A critical question in evaluating zero-shot VLN methods is determining how much performance gain originates from the foundation model (e.g., GPT-5) versus the navigation framework itself. To provide a comprehensive answer, we conducted a controlled comparative analysis involving four settings, as detailed in Table~\ref{tab:comprehensive_comparison}.

\paragraph{Baselines and Variants.} We compare the following configurations:
\begin{itemize}
    \item \textbf{AO-Planner (Original)}: The reported performance of the baseline using its default VLM~\cite{chen2025affordances}.
    \item \textbf{AO-Planner (GPT-5)}: Our re-implementation of AO-Planner using the exact same GPT-5 API as our method, evaluated on the 300-episode subset.
    \item \textbf{HSGM (w/o BEV Map)}: Our ablation variant where the VLM relies solely on egocentric visual prompts (similar to AO-Planner's input) without the top-down BEV map representation.
    \item \textbf{HSGM (Full Model)}: Our complete framework incorporating the Hierarchical Semantic-Geometric Map. We evaluate this setting using both \textbf{GPT-5} and \textbf{GPT-4o} backbones.
\end{itemize}

\paragraph{Results and Analysis.} 
The results indicate that while upgrading AO-Planner to GPT-5 improves its Success Rate to 32.3\%, it still significantly lags behind our method. 
The most critical comparison lies between \textbf{AO-Planner (GPT-5)} and our variant \textbf{HSGM (w/o BEV Map)}. Despite both utilizing the same backbone and egocentric inputs, our decoupled framework achieves a Success Rate of \textbf{46.0\%}, outperforming the upgraded AO-Planner by a substantial margin of \textbf{13.7\%}. This disparity exposes fundamental flaws in the visual prompting paradigm. First, regarding \textit{perception reliability}, we observed that AO-Planner's 2D segmentation (Grounded SAM) often hallucinates navigable areas on vertical surfaces (e.g., walls) due to texture similarities, causing the VLM to plan collision-prone paths. In contrast, HSGM employs 3D geometric constraints derived from depth data to physically enforce obstacle avoidance. Second, concerning the \textit{planning domain}, AO-Planner forces the VLM to infer 3D spatial dynamics implicitly from 2D pixels. Our method resolves this by fully decoupling reasoning from execution: the VLM solely identifies high-level waypoints, while the robust $A^*$ algorithm ensures precise low-level control. Thirdly, incorporating the global BEV Map in our Full Model further extends the lead to \textbf{51.0\%} SR, confirming the additional value of explicit global spatial modeling. Finally, when substituting the backbone with a less capable model(GPT-4o), HSGM still achieves a \textbf{41.3\%} SR.

\begin{table}[t]
    \centering
    \scriptsize
    \setlength{\tabcolsep}{2pt}
    \caption{\textbf{Latency Analysis.} \textbf{Top:} ID: Instr. Decompose, IS: Inst. Seg., PU: PCD Update, WG: Waypoint Gen., BR: BEV Raster., VQ: VLM Query (GPT-5), PP: Path Plan.}
    \label{tab:latency}
    
    \resizebox{\linewidth}{!}{%
    \begin{tabular}{lccccccc}
    \toprule
    \textbf{Module} & \textbf{ID} & \textbf{IS} & \textbf{PU} & \textbf{WG} & \textbf{BR} & \textbf{VQ} & \textbf{PP} \\
    \midrule
    \textbf{Latency} & 3.1s & 27ms & 165ms & 130ms & 77ms & 23.8s & 11ms \\
    \bottomrule
    \end{tabular}%
    }
    
    \vspace{0.8em} 
    
    \resizebox{\linewidth}{!}{%
    \begin{tabular}{lcccccc}
    \toprule
    \multirow{2}{*}{\textbf{Settings}} & \multicolumn{4}{c}{\textbf{HSGM (Ours)}} & \multicolumn{2}{c}{\textbf{AO-Planner~\cite{chen2025affordances}}} \\
    \cmidrule(lr){2-5} \cmidrule(lr){6-7} 
    & Dec. & Follow & Step Avg. & Episode & Step Avg. & Episode \\
    \midrule
    \textbf{Latency} & 24.3s & 192ms & \textbf{4.87s} & \textbf{341s} & 8.68s & 895s \\
    \bottomrule
    \end{tabular}%
    }
\end{table}

\begin{figure*}[htb]
    \centering
    \includegraphics[width=\linewidth]{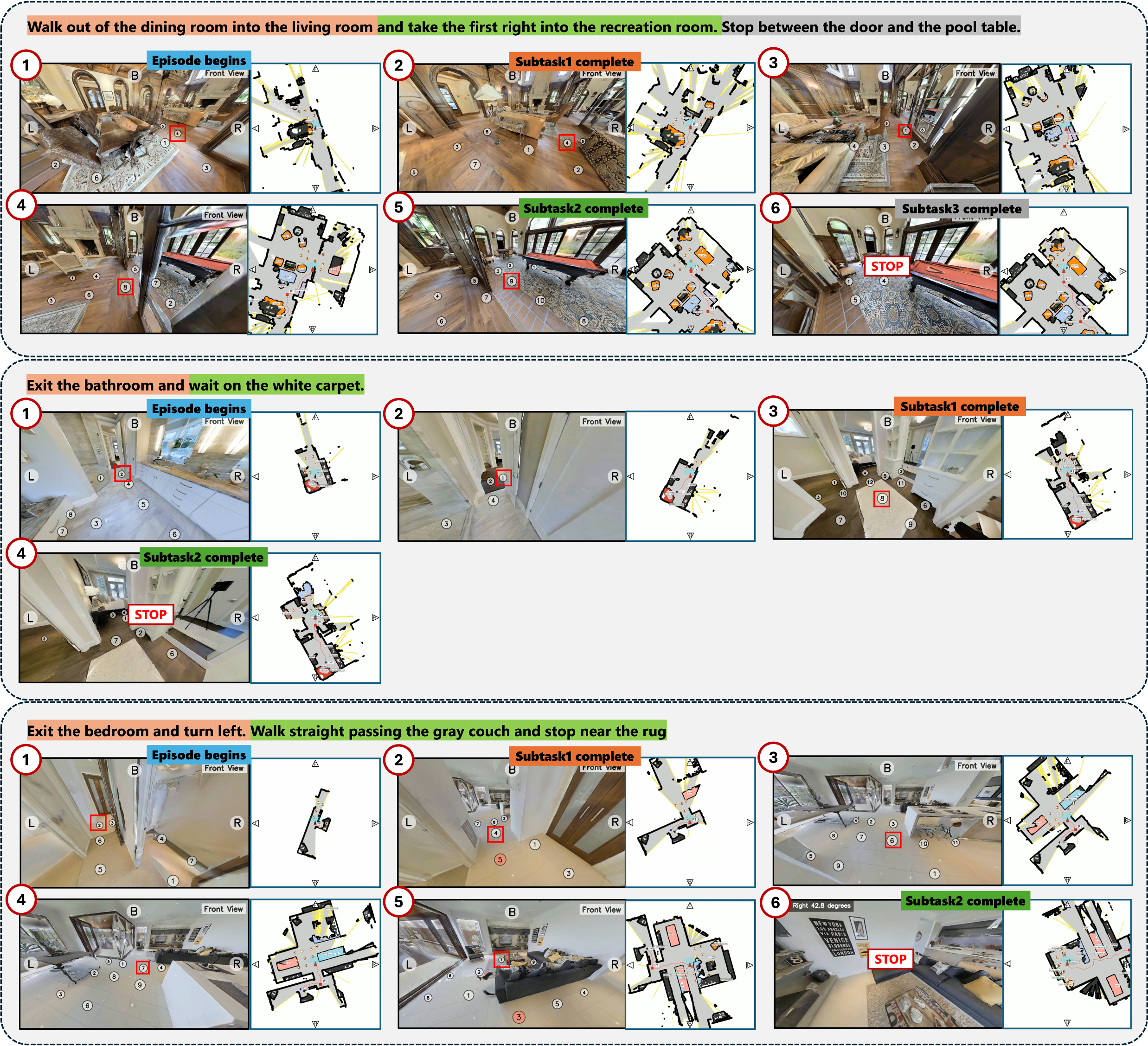}
    \caption{\textbf{Visualization of Success Cases.} We showcase three episodes demonstrating the agent's capability in multi-room traversal, object-referenced navigation (e.g., ``passing the gray couch''), and precise destination identification (e.g., ``wait on the white carpet'').}
    \label{fig:success_all}
\end{figure*}

\subsection{Latency and Token Cost}
\label{sec:latency_token_analysis}

We analyze system latency and VLM token consumption to evaluate efficiency. Agent operations are divided into \textit{Decision steps} (all modules invoked) and \textit{Path Following steps} (only perception and map updates). As shown in Table~\ref{tab:latency}, while Decision steps take 24.3s (dominated by the 23.8s VLM query), Path Following steps are highly efficient (192ms). Since multiple fast following steps occur between decisions, the overall average step latency is significantly amortized to \textbf{4.87s}. Compared to AO-Planner~\cite{chen2025affordances}, HSGM achieves superior navigation performance with nearly half the average step latency (4.87s vs. 8.68s) and a drastically shorter total episode time (\textbf{341s} vs. 895s).
In terms of token consumption, HSGM maintains a moderate level, with 319 tokens per decision and 4,229 tokens per episode.

\begin{figure*}[htb]
    \centering
    \includegraphics[width=\linewidth]{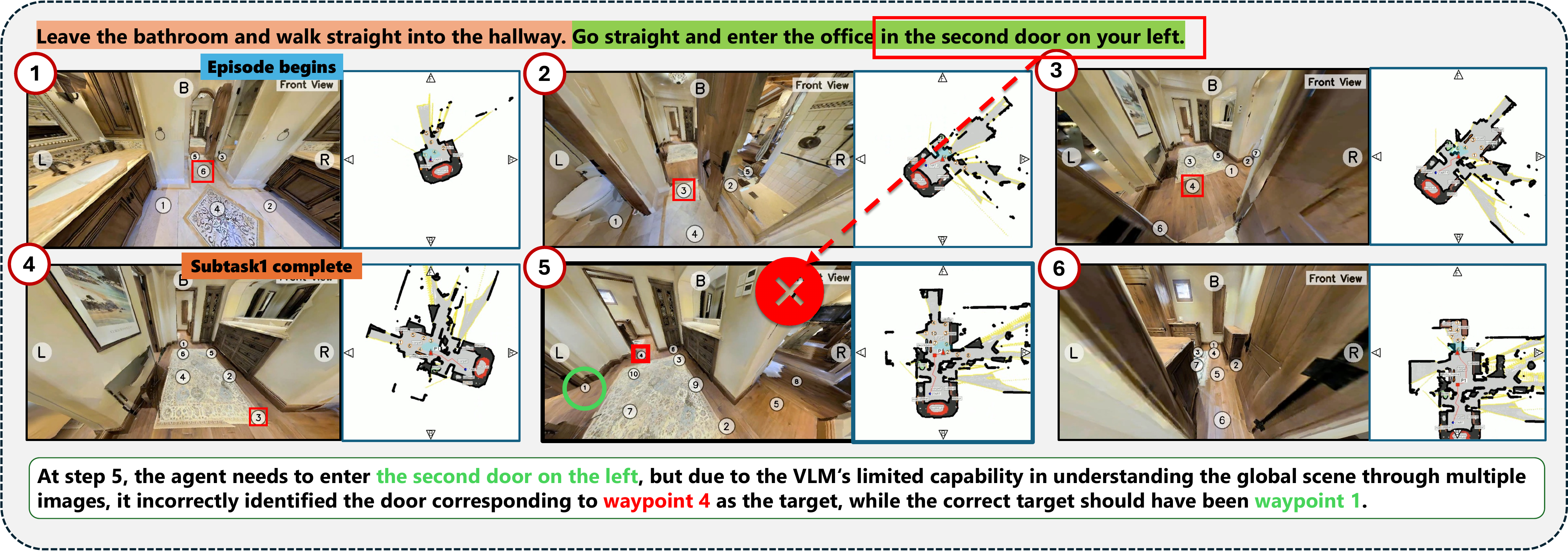}
    \caption{\textbf{Failure Case 1: Sequential Counting Error.}}
    \label{fig:fail1}
\end{figure*}

\begin{figure*}[htb]
    \centering
    \includegraphics[width=\linewidth]{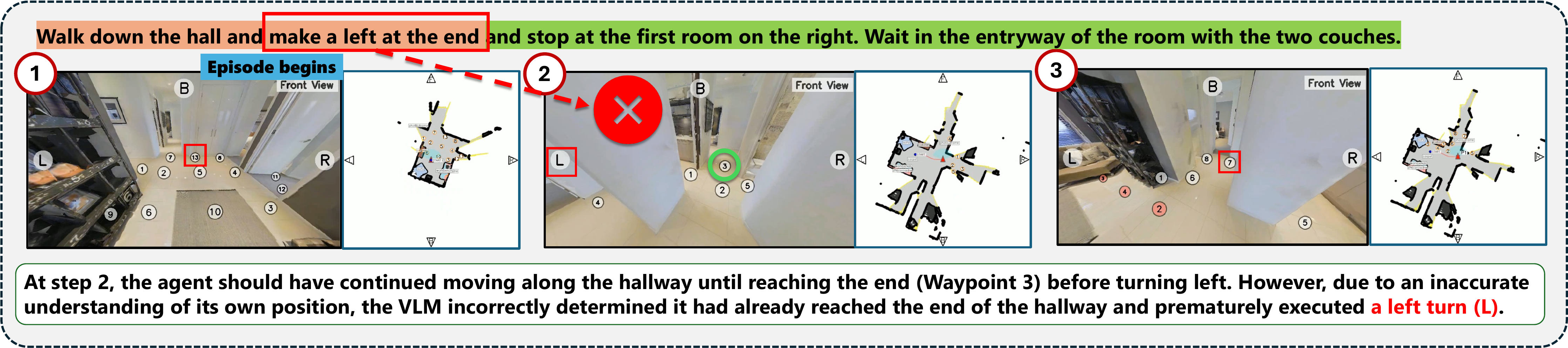}
    \caption{\textbf{Failure Case 2: Premature Execution.}}
    \label{fig:fail2}
\end{figure*}

\section{More Visualization Results}
\label{sec:more_vis}

\subsection{Success Cases}
\label{ssec:success}

As illustrated in Figure~\ref{fig:success_all}, our framework demonstrates robust performance across diverse and complex indoor environments. The visualization underscores the system's ability to decompose complex natural language instructions into manageable sequential subtasks, providing a clear and structured roadmap for long-horizon navigation. By synergizing global context from the BEV Map with local waypoint visual prompts, the VLM acts as an intuitive high-level planner, making reliable decisions to select geometrically valid targets.

Critically, our decoupled architecture ensures that these high-level semantic decisions are translated into precise physical actions. Once a target waypoint is selected, the underlying algorithm plans an optimal, collision-free path using $A^*$, enabling the agent to safely traverse cluttered environments that typically challenge end-to-end models. Furthermore, upon reaching the designated location for a specific subtask, the VLM effectively verifies the completion status, ensuring smooth transitions between subtasks or precise termination of the episode.

\subsection{Failure Case Analysis}
\label{ssec:failure}

Despite achieving state-of-the-art zero-shot performance, our qualitative analysis reveals specific limitations in the spatial reasoning capabilities of current VLMs, particularly regarding global scene understanding and precise self-localization.

The first failure mode, as shown in Figure~\ref{fig:fail1}, involves errors in identifying sequential landmarks due to fragmented global perception. In this episode, the agent is instructed to enter the ``second door on the left.'' However, at Step 5, the VLM fails to correctly interpret the global scene structure across multiple egocentric images. Instead of identifying the correct target (Waypoint 1), it incorrectly identifies the door corresponding to Waypoint 4 as the target. This misidentification highlights the VLM's limited capability in stitching together temporal observations to form a coherent global scene understanding, leading to failures in tasks requiring sequential counting.

The second failure mode, illustrated in Figure~\ref{fig:fail2}, pertains to the premature execution of actions caused by inaccurate state estimation. The instruction explicitly requires the agent to move along the hallway until reaching the end (Waypoint 3) before turning. However, at Step 2, due to an inaccurate understanding of its own position relative to the corridor's geometry, the VLM incorrectly determines that it has already reached the end of the hallway. Consequently, it prematurely executes a left turn (L) at an intermediate junction. This failure suggests that while the HSGM provides geometric layout, the VLM occasionally struggles to ground strict locational constraints (e.g., ``at the end'') against its current spatial state, prioritizing immediate directional affordances over geometric termination conditions.


\end{document}